\documentclass[letterpaper]{article}

\usepackage[top=1in, bottom=1in, left=0.8in, right=0.8in]{geometry}
\usepackage{graphicx}
\usepackage{amsmath}
\usepackage{amssymb}
\usepackage{subcaption} 

\newcommand{\norm}[1]{\left\lVert#1\right\rVert}

\usepackage[breaklinks=true,bookmarks=false]{hyperref}
\usepackage{breakurl}

\usepackage{hyperref} 
\usepackage{url} 
\usepackage{caption}
\usepackage{subcaption} 
\usepackage{graphicx} 

\begin{document}

\title{Qualitative Analysis of Monte Carlo Dropout}

\author{Ronald Seoh\\
University of Massachusetts Amherst\\
Amherst, MA\\
{\tt\small bseoh@cs.umass.edu}
}

\date{13 December 2019}

\maketitle

\begin{abstract}
In this report, we present qualitative analysis of Monte Carlo (MC) dropout method for measuring model uncertainty in neural network (NN) models. We first consider the sources of uncertainty in NNs, and briefly review Bayesian Neural Networks (BNN), the group of Bayesian approaches to tackle uncertainties in NNs. After presenting mathematical formulation of MC dropout, we proceed to suggesting potential benefits and associated costs for using MC dropout in typical NN models, with the results from our experiments.
\end{abstract}

\section{Introduction}
There is no doubt that rapid advances of neural network (NN) models in recent years have brought a profound impact in the field of machine learning and their applications in real-life systems. As we continue our efforts to apply NNs to even more complex and challenging problems, some have started to wonder about the risks involved with using them, especially in critical domains. Most of the widely used NN models today simply emits a single set of answers they were trained to provide. If life or death decision has to be made based on some predictions of these NNs, one would want to know how certain the models are about their outputs. 

Luckily, Bayesian modelling provides a framework for capturing model uncertainty - rather than coming up with a single set of model parameters, or simply the weights in NNs, we put a distribution around them and infer a posterior distribution of those weights given the data. This Bayesian way of modelling NN's uncertainty has generally been referred to as \emph{Bayesian Neural Network (BNN)} \cite{denker1991transforming}\cite{neal1995bayesian}. Despite being theoretically sound, it has been computationally challenging to do this in practice, especially with NNs getting deeper and being coupled with more complicated layers.

Gal and Ghahramani \cite{gal2016dropout} have introduced a shockingly simple method for capturing model uncertainty. They have discovered that training any NNs with \emph{dropouts}, typically used for preventing overfitting, could be interpreted as an approximate inference of the weight's posterior, as long as dropouts are added after every layer with weights. One simply needs to make multiple predictions with the trained model and average them. This method is now referred to as \emph{Monte Carlo (MC) dropout} in recent literatures.

Our project intended to provide further qualitative analysis on the benefits of MC dropout by empirically evaluating the \emph{practicality} of the method and suggesting potential benefits and costs when applying the method to actual models. While being very simple to implement, it is not easy to answer whether it would be appropriate for specific problems to make use of them, mainly because we first need to understand the nature of uncertainty information MC dropout captures. Plus, we would also have to consider whether potential increase in computational costs could negatively affect the ML system's effectiveness as a whole.

\section{Background}

\subsection{Uncertainty in NN models}
Before going into the analysis, it would be worthwhile to first explain what we mean by \emph{uncertainty} in neural networks. While the idea of defining uncertainty in any scientific modelling could be the subject of endless debate, most of the recent literatures on ML uncertainty seems to follow the categorization laid out by Der Kiureghian \& Ditlevsen \cite{der2009aleatory}, where they assert that uncertainties that arise from modelling largely falls into two categories:

\begin{enumerate}
\item \textbf{\emph{Epistemic uncertainty}} or \emph{model uncertainty}, which refers to the fact that we do not know the model that best explains the given data. For NNs, this is uncertainty from not knowing the best values of weights in all the trainable layers. This is often referred to as \emph{reducible} uncertainty, because we can theoretically reduce this type of uncertainty by acquring more data.
\item \textbf{\emph{Aleatoric uncertainty}} or \emph{data uncertainty}, where the underlying data generation might not have fully captured the pieces of information that should have been included. As a result, there would be inherent variability in possible explanations of this data. This is also called \emph{irreducible} uncertainty, as uncaptured information cannot be recovered even with infinite amount of data.
\end{enumerate}

As we will see in the next section, most of the endeavors, including MC dropout, have concentrated on addressing epistemic uncertainty in NN models. However, there have been some notable efforts to tackle aleatoric uncertainty, which build upon recent developments in Bayesian neural networks. This will be briefly discussed in the last section.

\subsection{Bayesian Neural Network}
Given the dataset \(D = (X, Y)\), where \(X = \{ x_1, x_2, ... , x_n \}\) contains the records of all the predictor variables and \(Y = \{ y_1, y_2, ... , y_n \}\) carries the response variable, we would like to predict new \(y^\star\) given some new data point \(x^\star\), with some NN model defined by the set of layer weights or \emph{parameters} \(W\). We typically formulate the task of finding optimal \(W\) as an optimization problem and solve it using techniques such as stochastic gradient descent. As a result, we get a single set of \(W\) and hence a single prediction of \(y^\star\).

In a nutshell, Bayesian Neural Network differs from the setting above as we try to model the distribution of \(W\), and subsequently the predictive posterior distribution of \(y^\star\).

\begin{align}
&P(D) = \int P(D \mid W) P(W) dW \\
&P(W \mid D) = \frac{P(D)P(W)}{P(D)} \\
&P(y^\star \mid x^\star, D) = \int P(y^\star \mid x^\star, W) P(W \mid D) dW \label{eq:bnn_predictive}
\end{align}

In fact, these are not really specific to BNNs - it's just Bayesian inference. However, the problem here is that the integral for \(P(D)\) is computationally feasible for very few cases of neural networks\footnote{It is known that a 1-layer fully-connected NN with i.i.d. prior over its parameters is a Gaussian Process (GP).\cite{neal1996priors}\cite{lee2017deep}} and consequently the integral for the posterior distribution of \(y^\star\) as well. 

To tackle cases like this where exact inference is difficult, Bayesians have developed two major families of appproaches to do \emph{approximate} inference. One is Variational Inference (VI) and another is Markov Chain Monte Carlo (MCMC). Hence, most of the historic and current literatures on BNNs have dealt with how we could do better approximate inference of the NN parameters using either of the two approaches.

We will provide very brief description of variational inference, of which MC dropout can be explained as one of its variants. For more detailed treatment of both VI and MCMC, please refer to the section 2.1 of \cite{gal2016uncertainty}, \cite{gal2015dropout}, \cite{Blei_2017}, and \cite{salimans2014markov}.

\subsection{Variational Inference (VI)}

First key idea in variational inference is that instead of sticking with \(P(W \mid D)\) in \autoref{eq:bnn_predictive}, we replace it with an approximate variational distribution \(q_\theta(W)\), which is parameterized by \(\theta\) and can be evaluated. Then we can rewrite \autoref{eq:bnn_predictive} as

\begin{align}
q_\theta^\star(D) &= \int P(y^\star \mid x^\star, W) q_\theta(W) dW \label{eq:posterior_vi} \\
                  &\approx P(y^\star \mid x^\star, D) \nonumber
\end{align}

We would want to make this \(q_\theta(W)\) to be closely resembling \(P(W \mid D)\) as possible. Ideally, we would do so by mimimizing the Kullbeck-Leibler (KL) divergence between the two:

\begin{align}
KL(q_\theta(W) \mid (P(W \mid D)) = \int q_\theta(W) \log \frac{q_\theta(W)}{P(W \mid D)} dW \label{eq:KL}
\end{align}

We cannot minimize \autoref{eq:KL} as it contains \(P(W \mid D)\). However, it has been found that we can minimize \autoref{eq:KL} by \emph{maximizing} evidence lower bound (ELBO) instead \cite{bishop2006pattern}:

\begin{align}
\mathcal{L}_{VI}(\theta) &= \int q_\theta(w) \log P(D \mid W) dW - KL(q_\theta(W) \mid P(W)) \label{eq:ELBO} \\
                       &\leq log P(D) \nonumber
\end{align}

The first term in \autoref{eq:ELBO} represents the conditional log-likelihood - that is, we want to make \(q_\theta(w)\) to explain \(D\) well by maximizing this term. The second term makes \(q_\theta(W)\) to be close to prior \(P(W)\). This essentially prevents the approximate distribution from being too complex and overfitted to the data.

Successful usage of VI largely depends on the choice of variational distribution, as evaluating \autoref{eq:ELBO} analytically might not be possible for more expressive distributions, while simpler distributions could create significant bias. Moreover, evaluating \autoref{eq:ELBO} naively would need to be done on the entire dataset, which could be a huge computational cost for large amounts of data. Those limitations kept the use of VI in BNNs, along with BNN itself, out of practical interests until fairly recently.

\subsubsection{MC dropout}

Monte Carlo dropout by Gal and Ghahramani \cite{gal2016dropout} can be explained as another way of performing variational inference on BNNs. We won't cover the full proof, but we will highlight few key details. 

In MC dropout, we define \(q_\theta (W_i) \), the approximating distribution for weights of each layer \(i\) to be

\begin{align}
&W_i = M_i \cdot diag( \lbrack Z_{i,j} \rbrack_{j=1}^{K_i} ) \\
&Z_{i,j} \sim \text{Bernoulli}(p_i) \text{ for } i = 1, ..., L, j = 1 ..., K_{i-1}.
\end{align}

where each \(Z_{i,j}\) is Bernoulli random variable deciding whether connected input should be dropped or not, with the probability \(p_i\). Hence \(p_i\) represents the probability of retaining the input, which is the exact opposite of what we know as \emph{dropout rate}.\footnote{\textbf{Note:} Some literatures consider the term `dropout rate' itself to be referring to the probability of keeping the input. However, we found this to be quite confusing given the English wording. We would like to make it clear that we will refer to `dropout rate' for the rest of this report as the rate of \emph{dropping} the input, not keeping, especially in line with how it is defined in popular deep learning frameworks such as \textsf{PyTorch} and \textsf{Tensorflow}.}

With this \(q_\theta (W)\), we would like to calculate \autoref{eq:ELBO}, but it is not possible to evaluate the first part with integral directly. Major breakthroughs in \cite{gal2016dropout} and \cite{gal2016uncertainty} can be summed up in two parts:

\begin{enumerate}
\item We could approximate that integral and get a unbiased estimator for \(\mathcal{L}_{VI}(\theta)\), by minimizing the typical loss functions with L2 regularization, typically used in learning NN models and takes the form of
    \begin{align}
    &\mathcal{L}_{\text{dropout}} = \frac{1}{N}\sum_{i=1}^{N} E(y_i, \hat{y_i}) + \lambda \sum_{i=1}^{L} (\norm{W_i}_2^2 + \norm{b_i}_2^2) \label{eq:typical_loss}
    \end{align}
where \(E(y_i, \hat{y_i})\) refers to loss functions such as softmax loss or mean squared error loss.
\item This variational inference is equivalent to a Gaussian Process (GP) approximation of neural network with model precision \(\tau\) and length scale \(l\), meaning we get an approximation of distribution over possible functions given data points.
\end{enumerate}

We can now get an approximate predictive posterior defined in \autoref{eq:bnn_predictive}, along with its mean and variance: 

    \begin{align}
    P(y^\star \mid x^\star, D) &= \mathcal{N}(y^\star; f^{W}(x^\star), \tau^{-1}\mathbf{I}) \label{eq:mc_dropout}\\
    \mathbb{E}_{q_\theta(y^\star \mid x^\star)}(y^\star) &\approx \frac{1}{T} \sum_{t=1}^{T} f^{W}(x^\star) \label{eq:mc_dropout_mean} \\
    \text{Var}_{q_\theta(y^\star \mid x^\star)}(y^\star) &\approx \tau^{-1} \mathbf{I}_D + \frac{1}{T} \sum_{t=1}^{T} (f^{W}(x^\star))^T f^{W}(x^\star) \nonumber \\
    &- (\mathbb{E}_{q_\theta(y^\star \mid x^\star)}(y^\star))^T\mathbb{E}_{q_\theta(y^\star \mid x^\star)}(y^\star) \label{eq:mc_dropout_var}
    \end{align}
where we are averaging \(T\) test predictions from the network trained with \autoref{eq:typical_loss}.

Please note that although MC dropout aims to capture model uncertainty, its formulation does not completely eliminate aleatoric uncertainty. Rather, as seen from \autoref{eq:mc_dropout}, it assumes homoscedastic aleatoric uncertainty, represented as \(\tau^{-1}\mathbf{I}\), indicating homogeneous data noise. There are efforts to model aleatoric uncertainty in different ways, which will be discussed in later sections.

The beauty is that we can do all this as if we are training typical neural network. Only differences are that we need to have dropout layers after every layers with weight parameters, and we need to make \(T\) test predictions.

\subsubsection{MC dropout hyperparameters}
\label{sec:mc_dropout_hyper}
It is important to note that MC dropout is a GP approximation with three hyperparameters:
\begin{itemize}
\item Retaining rate \(p\)
\item model precision \(\tau\)
\item length scale \(l\)
\end{itemize}

Model precision \(\tau\) is actually dependent of the other two. In \cite{gal2015dropout}, \(\tau\) is defined as

    \begin{align}
    &\tau = \frac{l^2p}{2N\lambda} \label{eq:tau}
    \end{align}
where \(p\) is the probability of retaining, \(\lambda\) is the regularization strength used in \autoref{eq:typical_loss}, and \(N\) is the length of training data. According to \cite{gal2016uncertainty}, setting length scale \(l\) can show our belief over function frequency - i.e. small (short) length-scale value meaning function output can change fairly rapidly, large (long) length-scale indicating values changing slowly. Without any strong assumptions for \(l\) yet, we can see that we could increase \(p\) in order to increase the overall model precision \(\tau\), at the expense of weaker regularization strengths \(\lambda\). Intuitively, this allows our approximate posterior to have higher predictive variance.

\subsubsection{In relation to standard dropout}

Some readers might wonder how MC dropout relates to the \emph{model averaging} effect of dropout. In fact, Srivastava et al. \cite{srivastava2014dropout} admits that more correct way of doing model averaging is to perform Monte Carlo sampling of different networks. However, They also empirically show that simply scaling the weights show comparable predictive performance in fully-connected networks, despite having no theoretical justification. \cite{gal2015bayesian} acknowledges this result, but argues that this is theoretically improper and MC dropout shows higher predictive performance in CNNs. In later sections, we confirm both results in our experiments. For the rest of this report, we will refer to this standard way of using dropout with weight scaling as \emph{standard dropout}, following term usage from recent literatures.

\section{Benefits}
With all the theoretical underpinnings of MC dropout, what can we make out of it? We'd like to highlight few key takeaways, backed by the results from our experiments.

\subsection{Uncertainty Information}
\label{sec:uncertainty_info}
First and foremost, the most obvious and essential benefit of MC dropout is the uncertainty information it captures.

In order to have more intuitive sense of uncertainty captured by MC dropout, we've decided to visualize the predictive distribution learned by MC dropout. We've created a synthetic dataset where we could clearly configure the hypothetical \emph{data generating function} and \emph{noise}. Similar to the setup used in Hernandez-Lobato and Adams \cite{hernandez2015probabilistic}, we created a dataset of 20 samples, with one predictor variable \(x\) randomly chosen from \([-4, 4]\), and response variable being \(y=x^3\), with noise of \(\epsilon \sim \mathcal{N}(0,\,9)\) added to \(y\). Then we trained a fully-connected 1-layer neural network, coupled with different types of non-linearity layers (ReLU, TanH) (these define different covariance functions for the GP) and the layer size of 100 with this toy dataset. With the trained network, we provide synthetic test data of same range with 100 points and see what MC dropout and standard dropout predicts.

\begin{figure}[h!]
\centering
\begin{subfigure}[t]{0.45\textwidth}
\includegraphics[width=\textwidth]{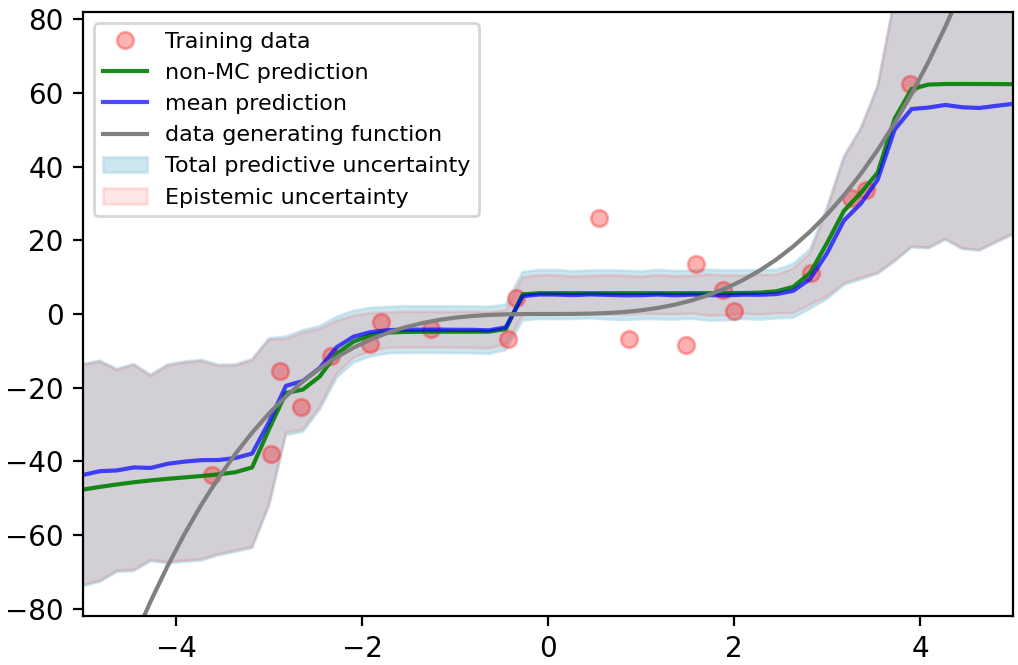}
\caption{TanH}
\label{fig:mc_dropout_tanh}
\end{subfigure}
\begin{subfigure}[t]{0.45\textwidth}
\includegraphics[width=\textwidth]{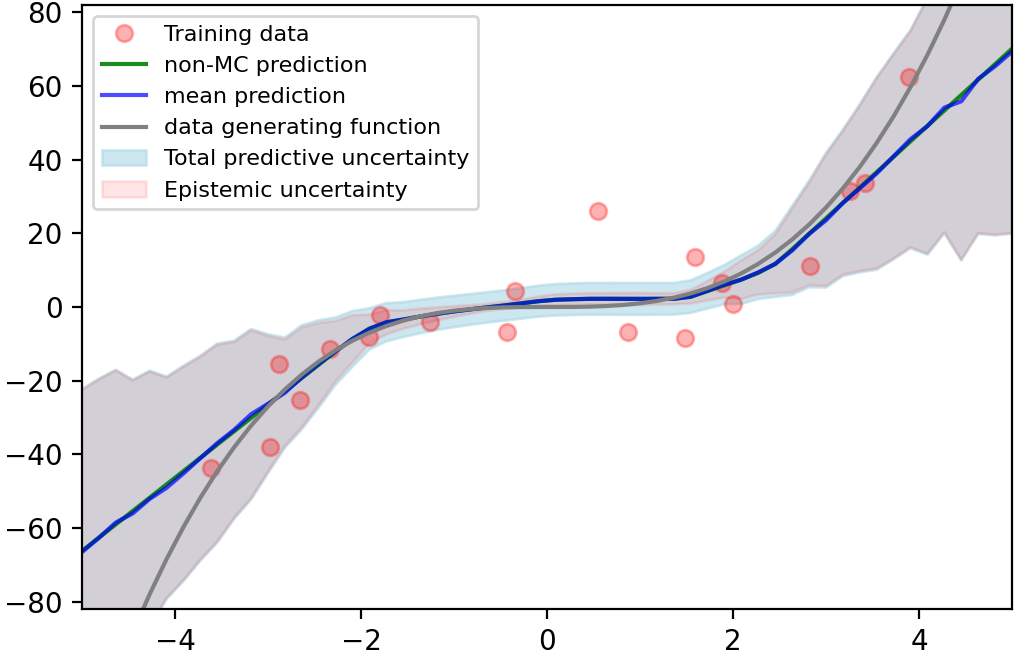}
\caption{ReLU}
\label{fig:mc_dropout_relu}
\end{subfigure}
\caption{Uncertainty captured from 1-layer fully-connected neural network, with different non-linearity layers. dropout rate 0.1, model precision 0.25.}
\label{fig:uncertainty_visual}
\end{figure}

The gray line is the data generating function, light blue line representing mean predictions by MC dropout, and light green line for predictions made by standard dropout. Red dots are the training data points. Blue shades, mostly overlapped with light red shades, represent areas 2 standard deviations away from each mean prediction. Within those blue shades, areas with red color show how much of them has been identified as epistemic uncertainty.

There are some interesting observations to be made from here. First of all, you can see thin blue colored areas wrapping red ones - as defined in \autoref{eq:mc_dropout_var}, that is homoscedastic noise of \(\tau^{-1}\) being added to epistemic uncertainty in the prediction of each points in the graph. Because we assume noise is homoscedastic, you can see that there are data points that fall outside the predictive variance measured. Secondly, all three have captured variance going up as we get further away from the center. This itself is not very surprising given the underlying data function we've used. However, they've done so in very different ways: \autoref{fig:mc_dropout_relu}'s uncertainty smoothly increases as we get to the both ends of the graph, but it seems that \autoref{fig:mc_dropout_tanh}'s uncertainty have more of staircase-like pattern: the amount of variance stays nearly constant for within different ranges of \(x\). Lastly, it is interesting to see how standard dropout largely matches MC dropout's predictions - it seems that this partially confirms \cite{srivastava2014dropout}'s claims of comparable performance. Especially for \autoref{fig:mc_dropout_relu}, standard dropout almost completely overlapped MC dropout, although it wasn't the case for \autoref{fig:mc_dropout_tanh}.

\subsubsection{Effects of dropout rates and model precision}

After looking at \autoref{fig:uncertainty_visual}, we also want to see the effects of changing dropout rates and model precision.

\begin{figure}[h!]
\centering
\begin{subfigure}[t]{0.225\textwidth}
\includegraphics[width=\textwidth]{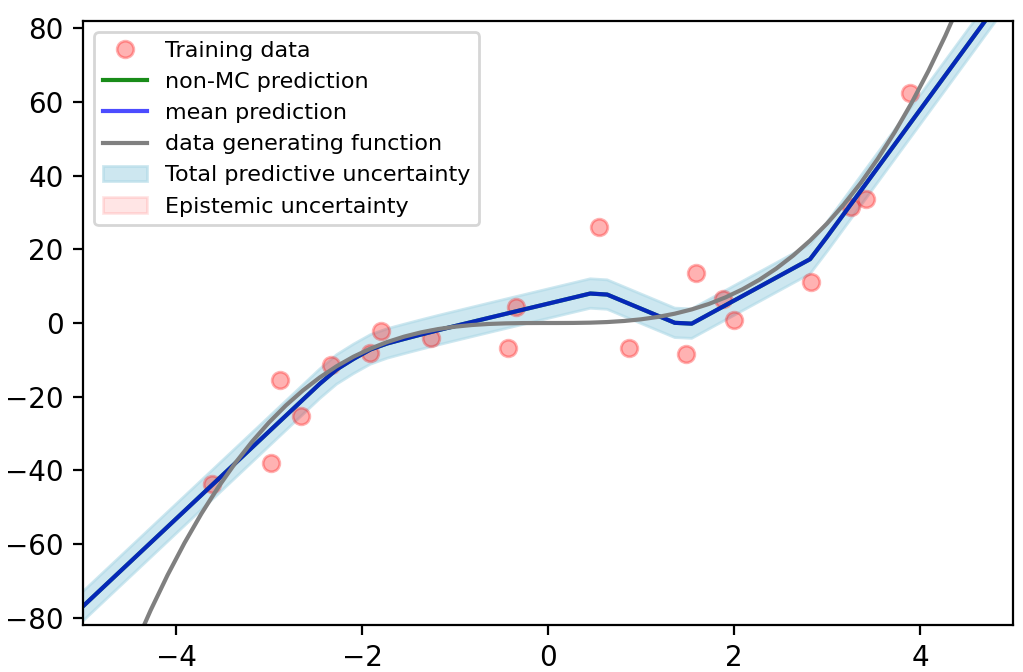}
\caption{0}
\label{fig:dropout_rate_0}
\end{subfigure}
\begin{subfigure}[t]{0.225\textwidth}
\includegraphics[width=\textwidth]{./mc_dropout_relu.png}
\caption{0.1}
\label{fig:dropout_rate_0_1}
\end{subfigure}
\begin{subfigure}[t]{0.225\textwidth}
\includegraphics[width=\textwidth]{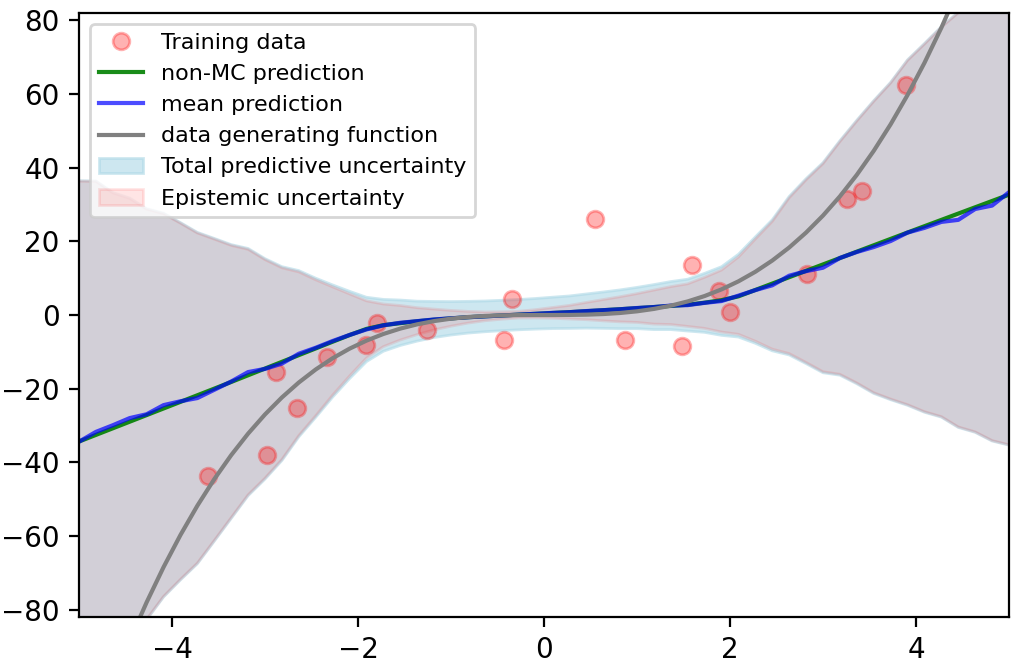}
\caption{0.5}
\label{fig:dropout_rate_0_5}
\end{subfigure}
\begin{subfigure}[t]{0.225\textwidth}
\includegraphics[width=\textwidth]{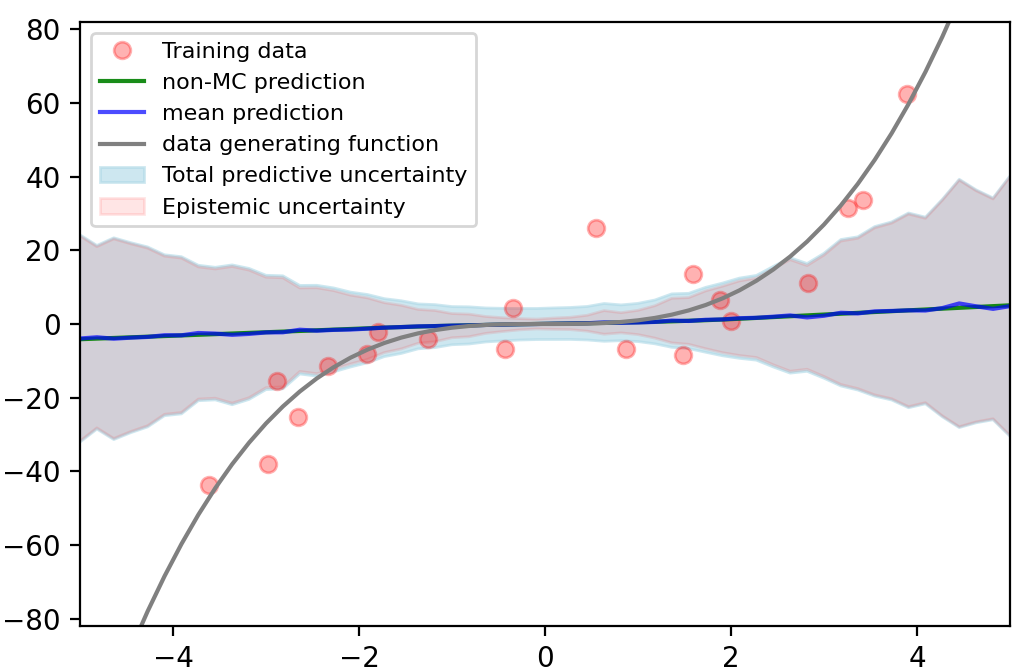}
\caption{0.9}
\label{fig:dropout_rate_0_9}
\end{subfigure}
\caption{MC dropout results on the toy dataset with different dropout rates. \(\tau\) fixed at 0.25. ReLU non-linearity used.}
\label{fig:uncertainty_visual_dropout}
\end{figure}

First with dropout rates, what's interesting is how mean predictions and red shades change as dropout rate goes up. When we look at \autoref{fig:dropout_rate_0}, you can see that our model almost completely overfitted into the underlying data function. This is not surprising given our test setup. However, as we increase dropout rate, not only our mean predictions start to diverge from the underlying function, but the red shaded area also gets larger. With the rate of 0.9, the mean prediction line became almost horizontal, although the shaded area got smaller.

Given our existing intuition for dropout, higher rates like \autoref{fig:dropout_rate_0_9} would mean that our model was regularized too much and learned too little of inherent patterns in the data. Under the MC dropout formulation, there's slightly more rigorous interpretation of this behavior, which starts from the fact that MC dropout is a Gaussian Process approximation. Since MC dropout is a variational inference method, we get to maximize the evidence lower bound (ELBO), which is the sum of likelihood function minus the KL divergence between the approximating distribution and prior distribution. The prior is defined as \(\mathcal{N}(0, \hat{\mathbf{K}}(\mathbf{X}, \mathbf{X}))\), where \(\hat{\mathbf{K}}\) is our estimated covariance function. By using approximating distribution of high dropout rate (low Bernoulli probability), we ended up minimizing the KL divergence from this prior distribution of mean 0, while having very low likelihood.

It might seem just foolish to use high dropout rate at this point, as no one would think that \autoref{fig:dropout_rate_0_9} is a good estimate of predictive uncertainty of our model. However, that's just because we know in advance what the underlying data function is. In real life, we cannot really identify the hypothesis space the true underlying function resides. We just simply assume that whatever model we use could be very close to that hypothesis space. 

Although it is beyond the scope of this project, one future avenue of research would be on whether we could make use of dropout rate to formally encode our belief towards how certain we are about training data containing information about the underlying data distribution. Given that dropouts has been used to prevent overfitting and not to learn too much from data, we feel that this view is somewhat implicit.

\begin{figure}[h!]
\centering
\begin{subfigure}[t]{0.325\textwidth}
\includegraphics[width=\textwidth]{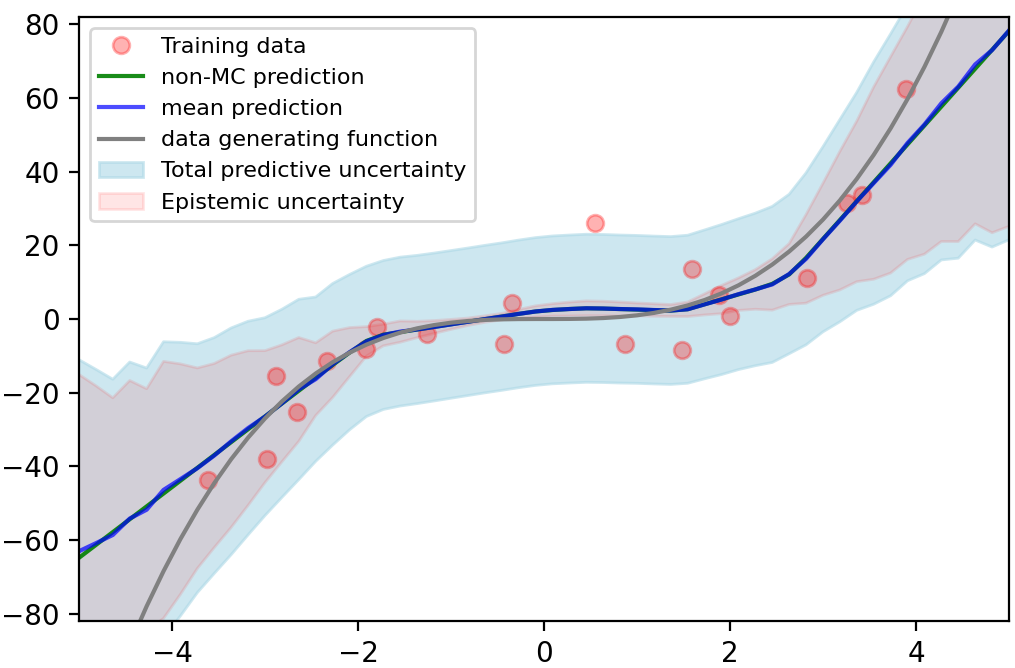}
\caption{model precision 0.01}
\label{fig:mc_dropout_relu_tau_0_01}
\end{subfigure}
\begin{subfigure}[t]{0.325\textwidth}
\includegraphics[width=\textwidth]{./mc_dropout_relu.png}
\caption{model precision 0.25}
\label{fig:mc_dropout_relu_tau_0_25}
\end{subfigure}
\begin{subfigure}[t]{0.325\textwidth}
\includegraphics[width=\textwidth]{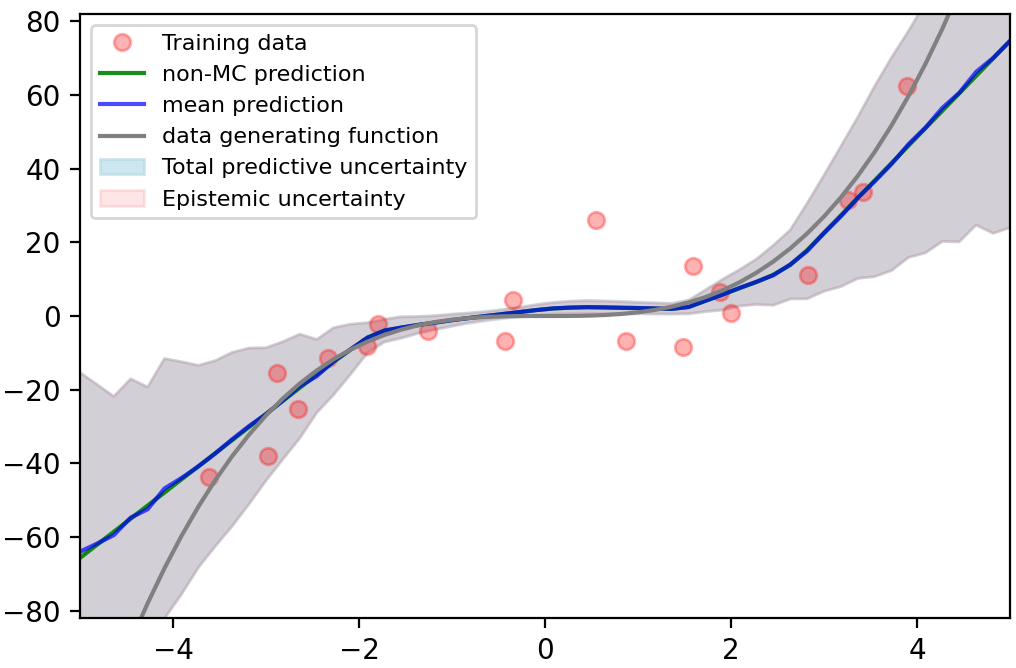}
\caption{model precision 10}
\label{fig:mc_dropout_relu_tau_10}
\end{subfigure}
\caption{Effects of different model precisions. dropout rate fixed at 0.1. ReLU non-linearity used.}
\label{fig:uncertainty_visual_tau}
\end{figure}

Unlike dropout rate, model precision \(\tau\) does not seem to change the overall trends of uncertainty captured, judging from \autoref{fig:uncertainty_visual_tau}. As previously mentioned, we add constant noise of \(\tau^{-1}\) throughout all the points of the predictions, and you can see that by the time we use \(\tau = 10\) in \autoref{fig:mc_dropout_relu_tau_10}, that noise is almost non-existent.

\subsubsection{Training Epochs}

Lastly, we wanted to visualize how much number of training epoch matters.

\begin{figure}[h!]
\centering
\begin{subfigure}[t]{0.325\textwidth}
\includegraphics[width=\textwidth]{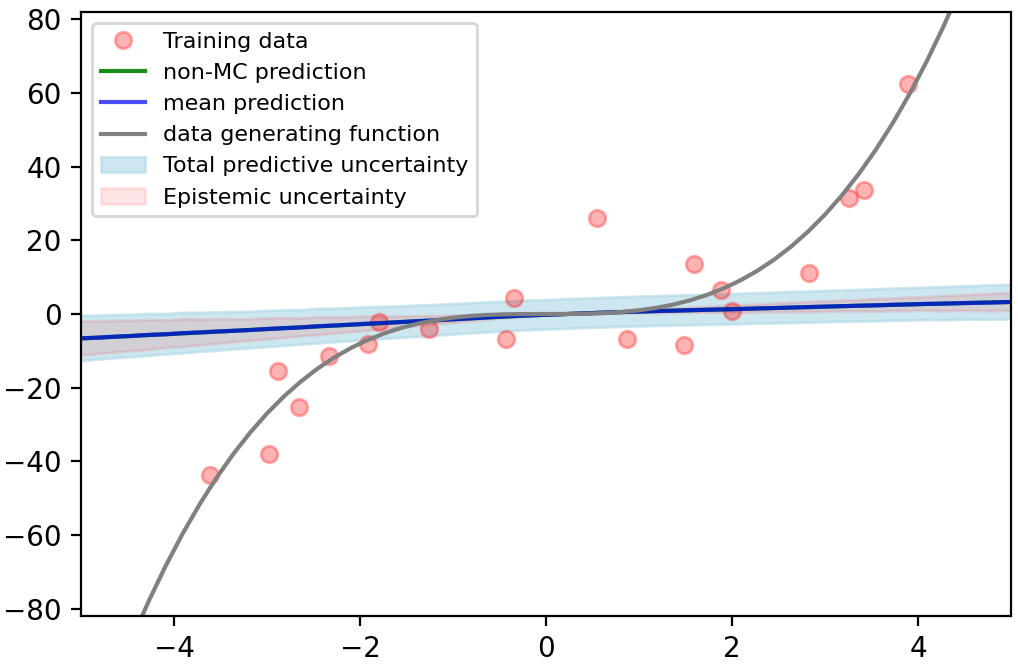}
\caption{40 epochs}
\label{fig:mc_dropout_relu_epochs_40}
\end{subfigure}
\begin{subfigure}[t]{0.325\textwidth}
\includegraphics[width=\textwidth]{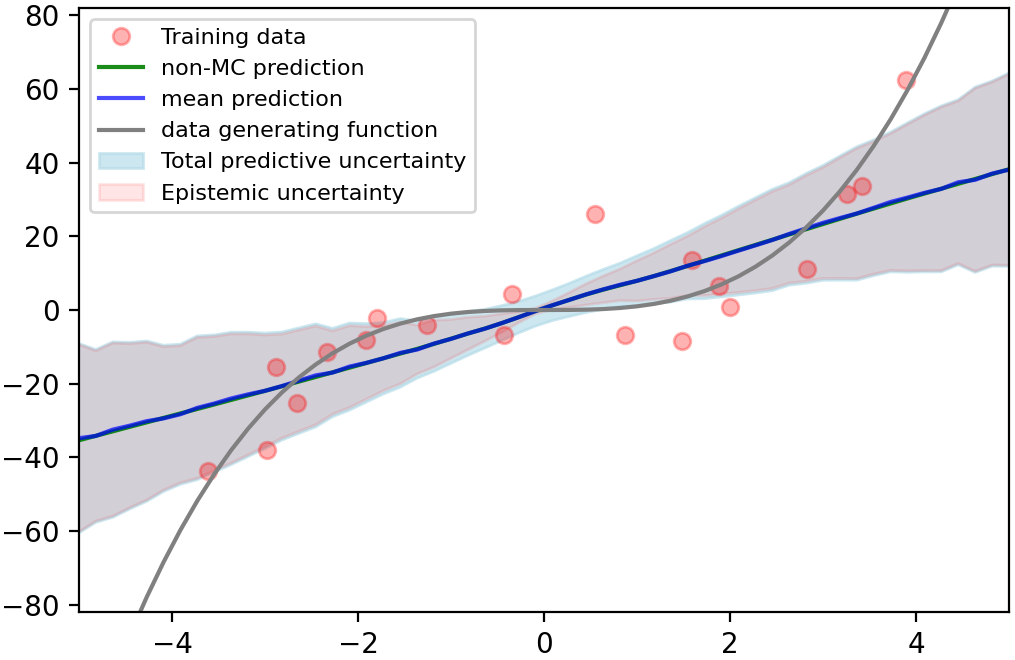}
\caption{400 epochs}
\label{fig:mc_dropout_relu_epochs_400}
\end{subfigure}
\begin{subfigure}[t]{0.325\textwidth}
\includegraphics[width=\textwidth]{./mc_dropout_relu.png}
\caption{4000 epochs}
\label{fig:mc_dropout_relu_epochs_4000}
\end{subfigure}
\caption{Effects of different model precisions. \(\tau = 0.25\), dropout rate \(= 0.1\), and ReLU non-linearity.}
\label{fig:uncertainty_visual_epochs}
\end{figure}

We can see that with 40 epochs, our NN underfitted, and captures nearly none of the supposed uncertainty around our model. Training 400 epochs somewhat improves situation but still does not capture relatively smaller uncertainty in the range where training data is given. We believe this intuitively explains what MC dropout is doing in terms of \autoref{eq:ELBO}, since we are trying to make our approximation of predictive posterior close as possible to our supposed GP posterior. At lower number of epochs, we could say that we are not yet close to resembling the true posterior.

\subsection{Improvements in Predictive Performance}

It is generally understood that standard dropouts improve perdictive performance of neural networks. We confirm that the improvement extends to MC dropout as well. Moreover, we wanted to see if MC dropout allows extra positive gains over standard dropout.

\begin{figure}[h!]
\centering
\begin{subfigure}[t]{0.45\textwidth}
\includegraphics[width=\textwidth]{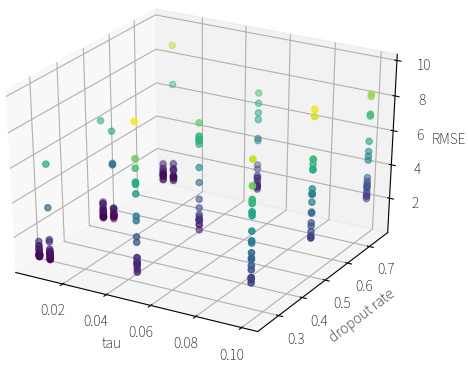}
\caption{Without dropout}
\label{fig:yacht_100_rmse}
\end{subfigure}
\begin{subfigure}[t]{0.45\textwidth}
\includegraphics[width=\textwidth]{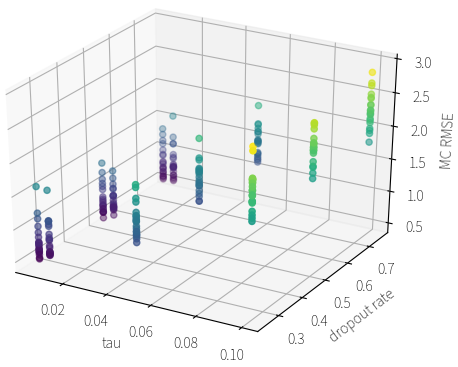}
\caption{With MC dropout}
\label{fig:yacht_100_MC_rmse}
\end{subfigure}
\caption{Spread of RMSE for regression task on \textsf{yacht} dataset.}
\label{fig:yacht_rmse}
\end{figure}

We trained a 1-layer neural network for regression task on \textsf{yacht} dataset, which only have 308 data points and one of the smallest dataset chosen for analysis. In \autoref{fig:yacht_rmse}, each point in all the vertical lines stands for validation results of the model with hyperparameter combinations, done for 20 different random train-test splits of the original dataset. You can see that MC dropout not only lowered RMSE error (Note different z-scales for two figures.), but variance across different splits. This was especially encouraging, given that we trained on very small amount of data. This could be the sign that we can make more robust predictions with NN, even when there's small amount of data.

\begin{figure}[p]
\centering
\includegraphics[width=0.76\textwidth]{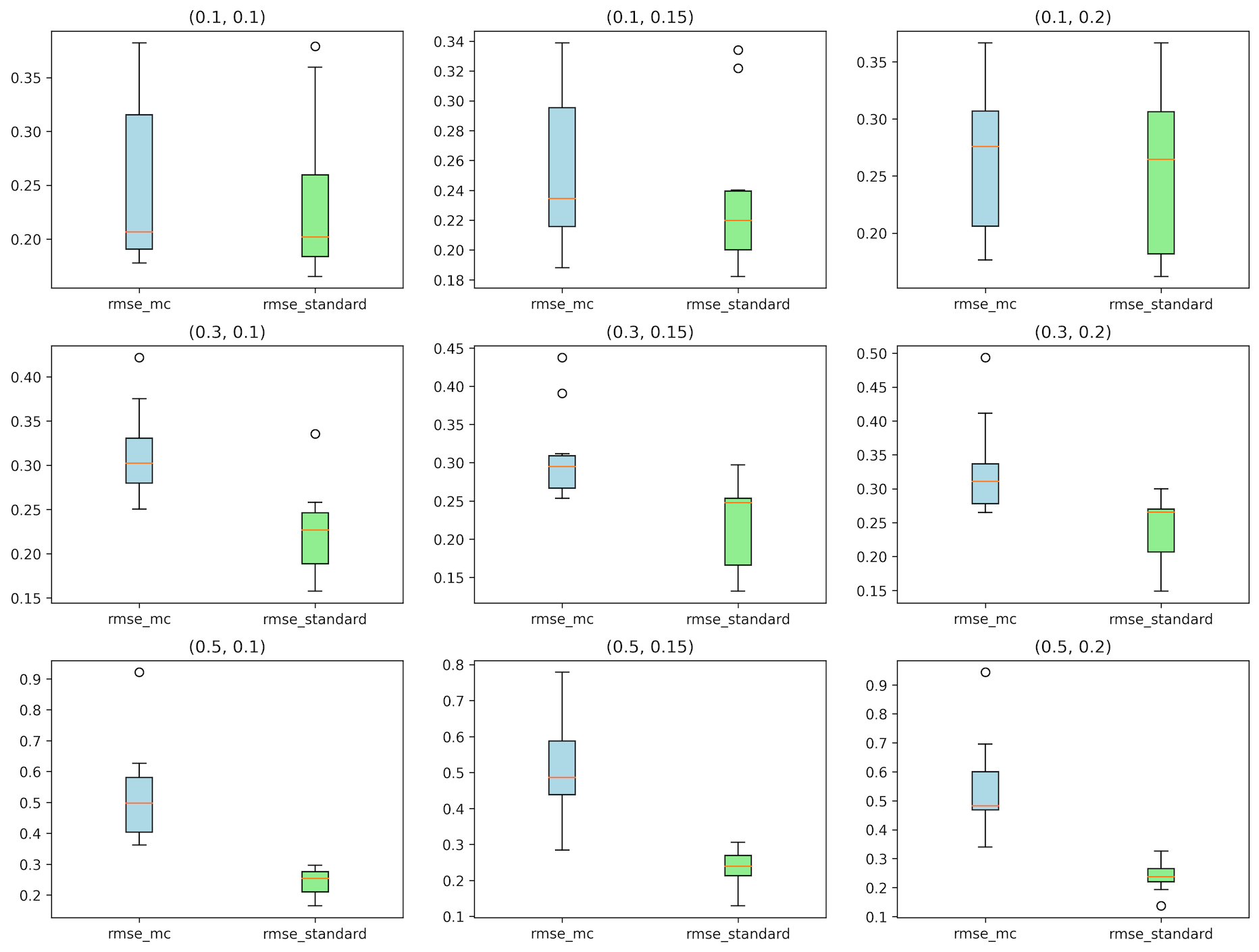}
\caption{Boxplots for RMSE scores obtained for different splits of \textsf{bostonHousing} dataset. Low values are better. X-axis represents different model precision \(\tau\) values and y-axis representing dropout rates.}
\label{fig:boxplot_bostonHousing}
\end{figure}

However, it was not entirely clear whether there are performance advantages over standard dropout. \autoref{fig:boxplot_bostonHousing} shows boxplots of RMSE scores across 10 train-test splits for MC dropout and standard dropout, done on a 1-layer FC NN and \textsf{bostonHousing} dataset (506 rows). It appears that MC dropout did not show superior error rates or more stable range of scores across splits in any cases. We tested other datasets as well and have seen similar trends. This also seems to be in line with our observation made earlier in \autoref{fig:mc_dropout_relu}.

\begin{figure}[p]
\centering
\includegraphics[width=0.76\textwidth]{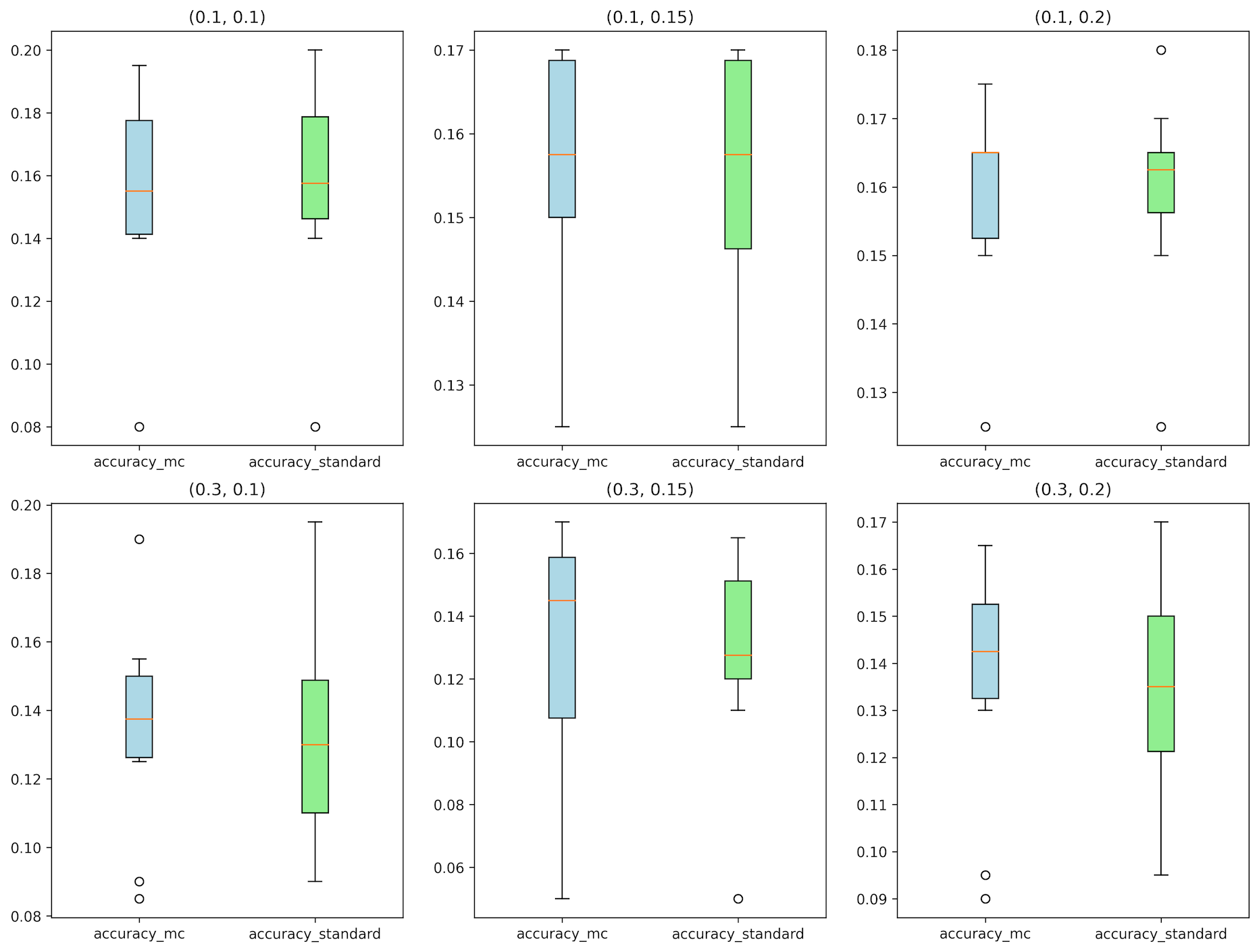}
\caption{Boxplots for prediction accuracies obtained by CNN with different splits of \textsf{CIFAR-10} dataset. \textbf{Higher} values are better. X-axis represents different model precision \(\tau\) values and y-axis representing dropout rates.}
\label{fig:boxplot_cifar10}
\end{figure}

\cite{gal2015bayesian} have shown that MC dropout does show significant improvement in convolutional neural networks. Hence, we wanted to check whether this holds consistently, especially when we have smaller number of data points. Due to issues in implementation and time limitations, we unfortunately were not able to fully test this claim. \autoref{fig:boxplot_cifar10} shows boxplots of accuracies obtained with CNNs for different splits of CIFAR-10. Note that we attempted to limit the number of data available to be just 1\% of the whole dataset, which is 500 points out of 50,000.

\section{Costs}
We now turn to discussing costs associated with MC dropout.

\subsection{Number of Training Epochs}

We have seen in \autoref{fig:uncertainty_visual_epochs} that MC dropout would relatively high number of epochs to get good uncertainty estimates. Early on in our project, we first attempted to replicate scores reported in \cite{gal2016dropout} by replicating its test setting. 

\begin{table}[!htbp]
  \centering
  \resizebox{17cm}{!}{
  \begin{tabular}{|c|c|c|c|c|c|}
  \hline
                                           & \textit{Size}   & RMSE (Dropout-Ron) & RMSE (Dropout-Gal) & LL (Dropout-Ron)   &  LL (Dropout-Gal) \\
  \hline
  \textsf{yacht}                           & \textit{308}    & 2.82 \(\pm\) 0.20  & 0.67 \(\pm\) 0.05  & -2.32 \(\pm\) 0.07 & -1.25 \(\pm\) 0.01 \\
  \hline
  \textsf{bostonHousing}                   & \textit{506}    & 3.08 \(\pm\) 0.19  & 2.90 \(\pm\) 0.18  & -2.88 \(\pm\) 0.11 & -2.40 \(\pm\) 0.04 \\
  \hline
  \textsf{power-plant}\(\star\)            & \textit{9,568}  & 4.19 \(\pm\) 0.03  & 4.01 \(\pm\) 0.04  & -2.84 \(\pm\) 0.01 & -2.80 \(\pm\) 0.01 \\
  \hline
  \textsf{naval-propulsion-plant}\(\star\) & \textit{11,934} & 0.01 \(\pm\) 0.00  & 0.00 \(\pm\) 0.00  &  3.46 \(\pm\) 0.01 &  4.45 \(\pm\) 0.00 \\
  \hline
  \end{tabular}
  }
    \caption{Reported RMSE, Log-likelihood, and standard errors for each. Lower RMSE scores / Higher LLs are better. `Dropout-Ron' is our result. \(\star\) denotes cases where we only ran one-tenth of original SGD iterations (epochs).}
  \label{tab:rmse_comparison}
\end{table}

\begin{table}[!htbp]
  \centering
  \begin{tabular}{|c|c|c|}
  \hline
              &            RMSE           &              LL            \\
  \hline
  40 Epochs   & 9.186061 \(\pm\) 0.486915 & -7.326624 \(\pm\) 0.773170 \\
  \hline
  400 Epochs  & 2.816304 \(\pm\) 0.196549 & -2.321702 \(\pm\) 0.069091 \\
  \hline
  4000 Epochs & 0.719461 \(\pm\) 0.052636 & -1.268644 \(\pm\) 0.019095 \\
  \hline
  \end{tabular}
    \caption{Comparison of scores for \textsf{yacht} dataset with different number of epochs.}
  \label{tab:epochs_comparison}
\end{table}

You can see that \autoref{tab:rmse_comparison} that our numbers do not quite match the scores reported by \cite{gal2016dropout}. We investigated this issue and found out that you need more number of SGD iterations to get similar performance, as seen in \autoref{tab:epochs_comparison}. Along with our qualitative analysis done earlier, we can see that one would need to train NN longer to get good uncertainty estimates.

\begin{figure}[h!]
\centering
\begin{subfigure}[t]{0.32\textwidth}
\includegraphics[width=\textwidth]{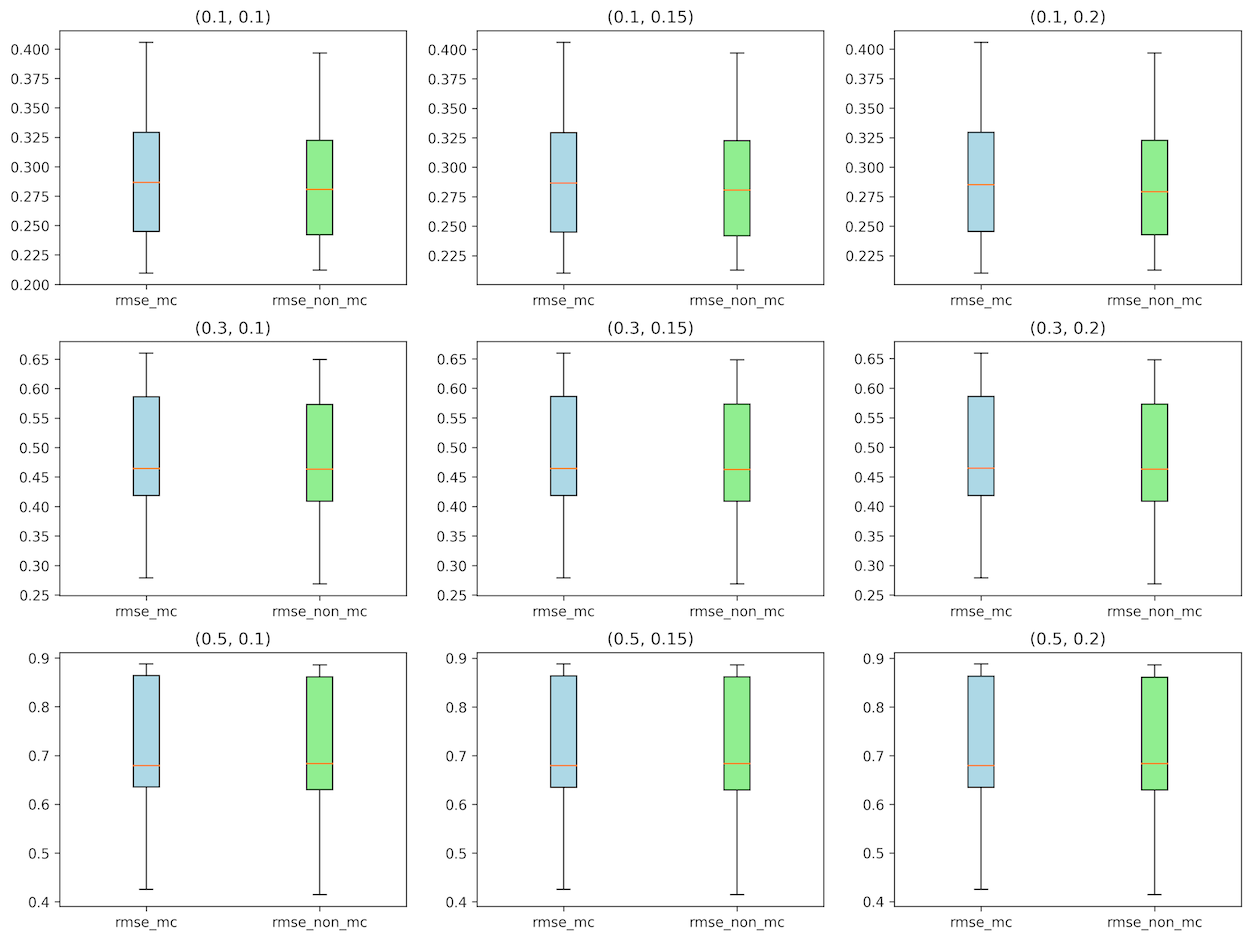}
\caption{1 layer, 40 epochs}
\label{fig:1_0.01_1_40box}
\end{subfigure}
\begin{subfigure}[t]{0.32\textwidth}
\includegraphics[width=\textwidth]{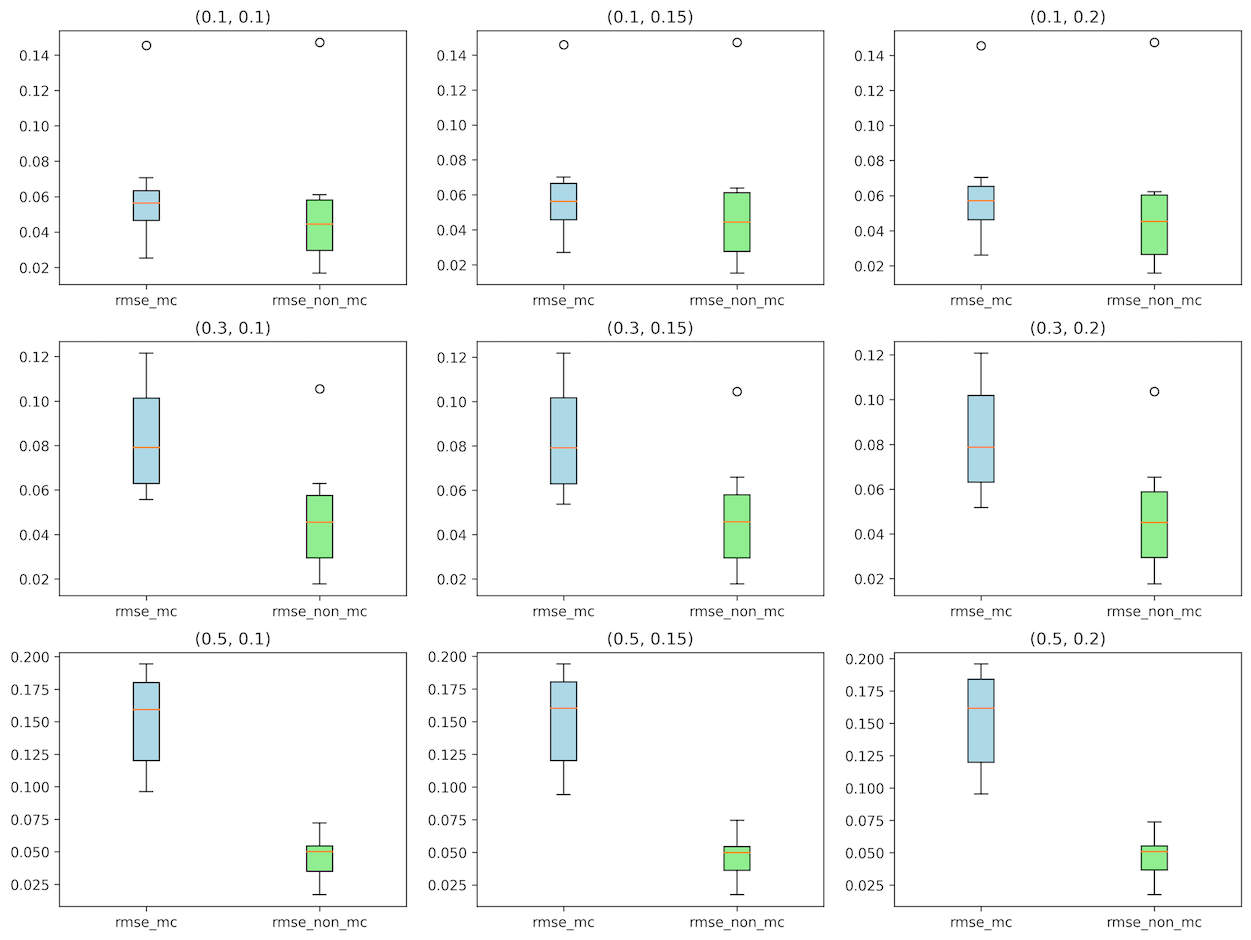}
\caption{1 layer, 400 epochs}
\label{fig:1_0.01_1_400box}
\end{subfigure}
\begin{subfigure}[t]{0.32\textwidth}
\includegraphics[width=\textwidth]{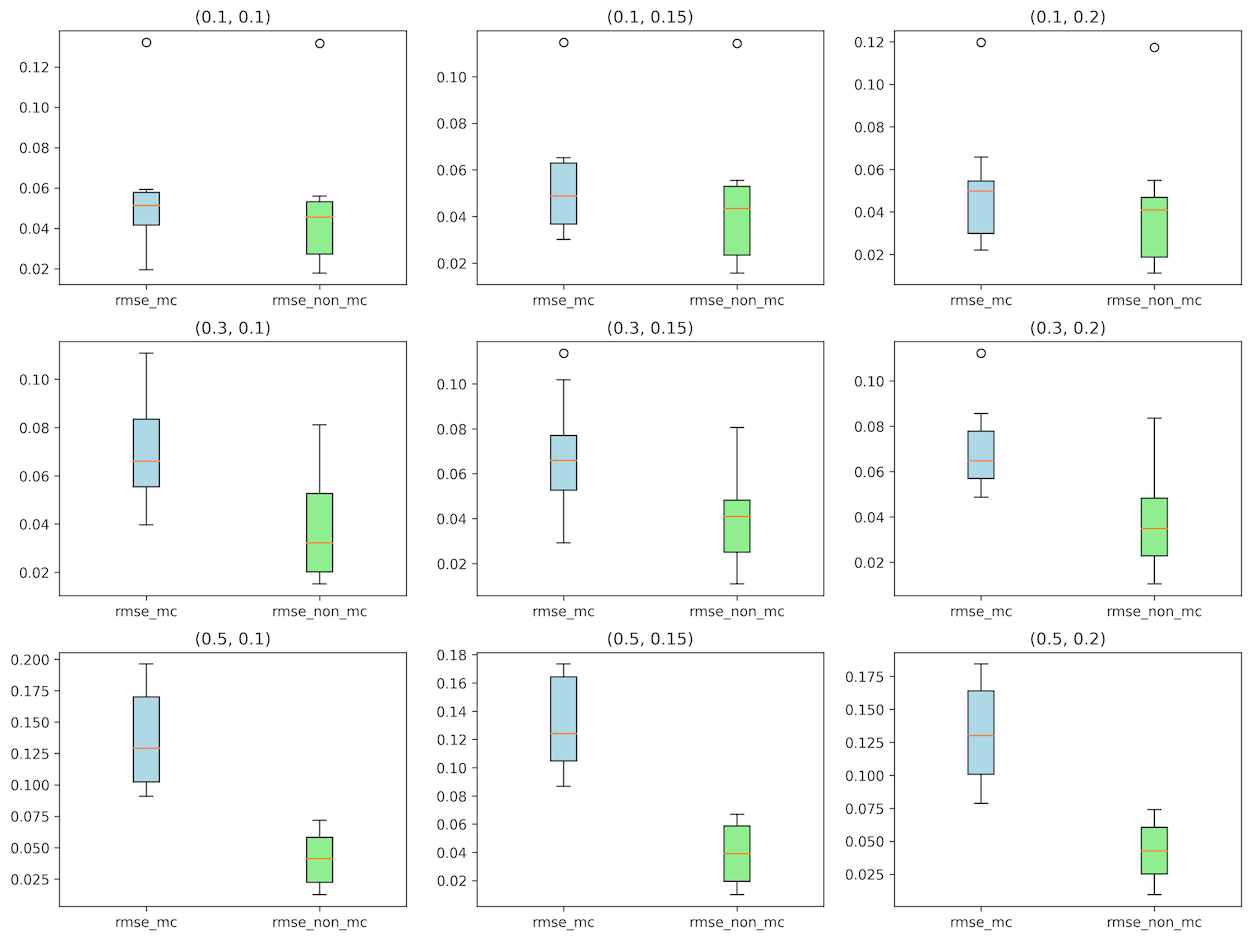}
\caption{1 layer, 4000 epochs}
\label{fig:1_0.01_1_4000box}
\end{subfigure}
\begin{subfigure}[t]{0.32\textwidth}
\includegraphics[width=\textwidth]{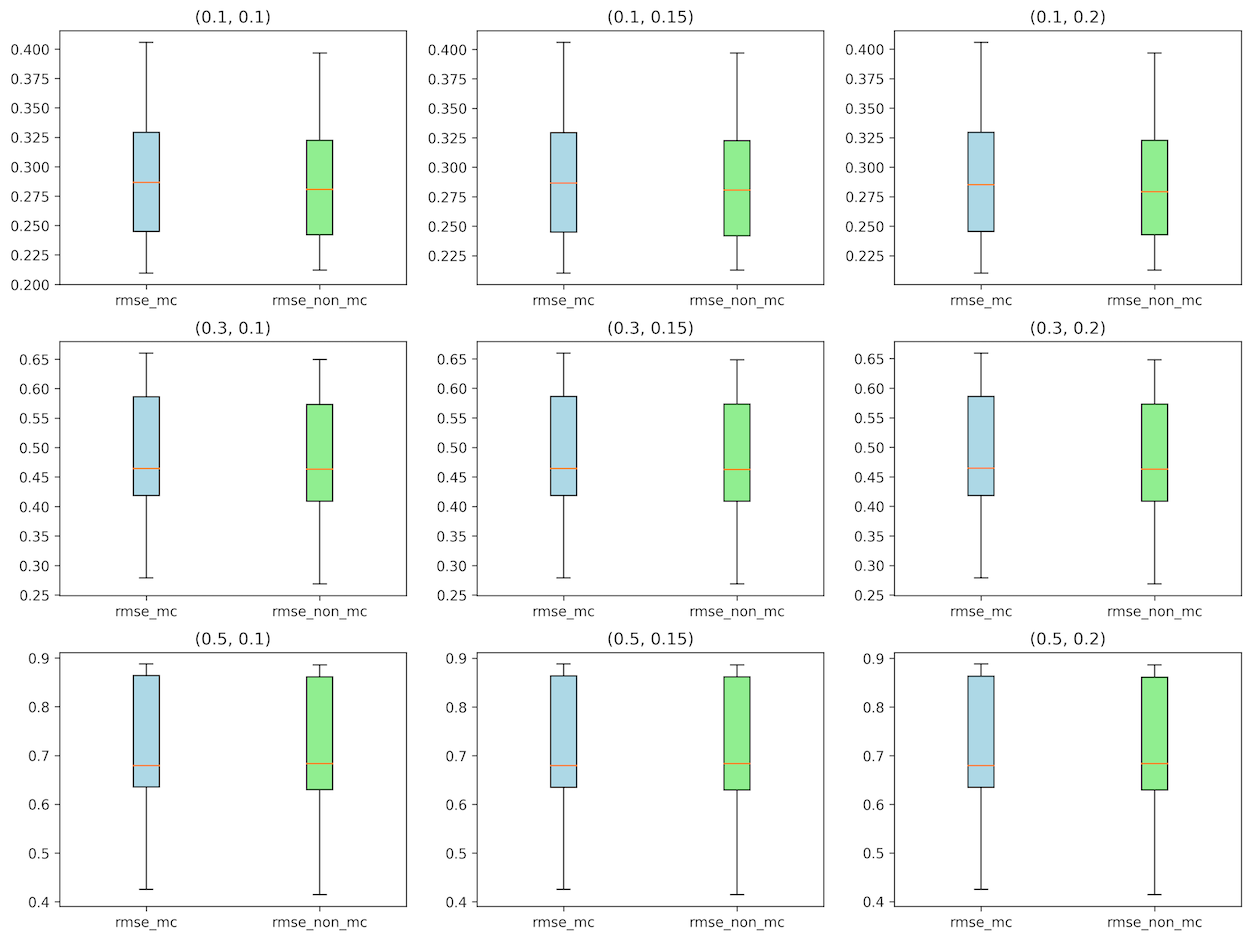}
\caption{3 layers, 40 epochs}
\label{fig:1_0.01_3_40box}
\end{subfigure}
\begin{subfigure}[t]{0.32\textwidth}
\includegraphics[width=\textwidth]{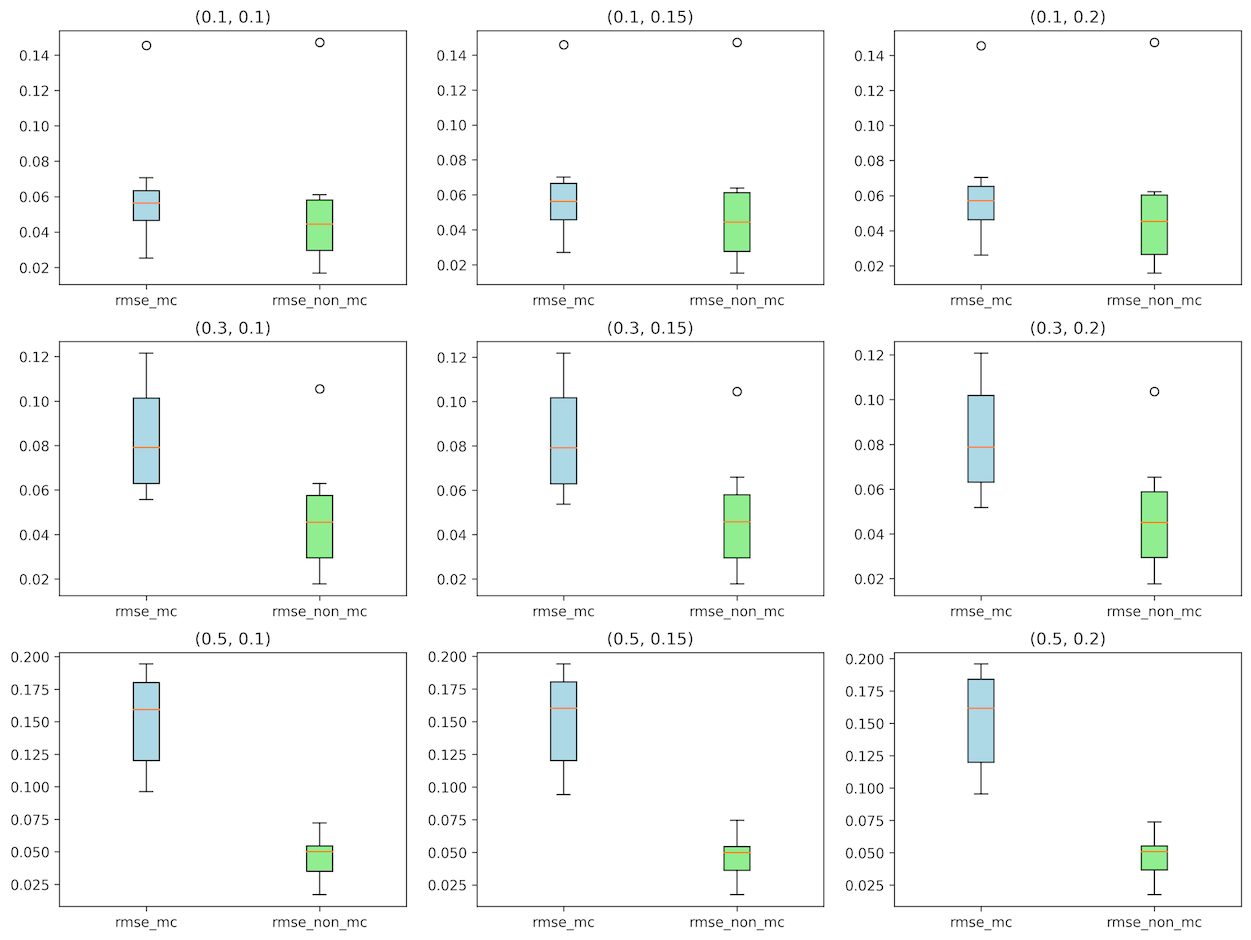}
\caption{3 layers, 400 epochs}
\label{fig:1_0.01_3_400box}
\end{subfigure}
\begin{subfigure}[t]{0.32\textwidth}
\includegraphics[width=\textwidth]{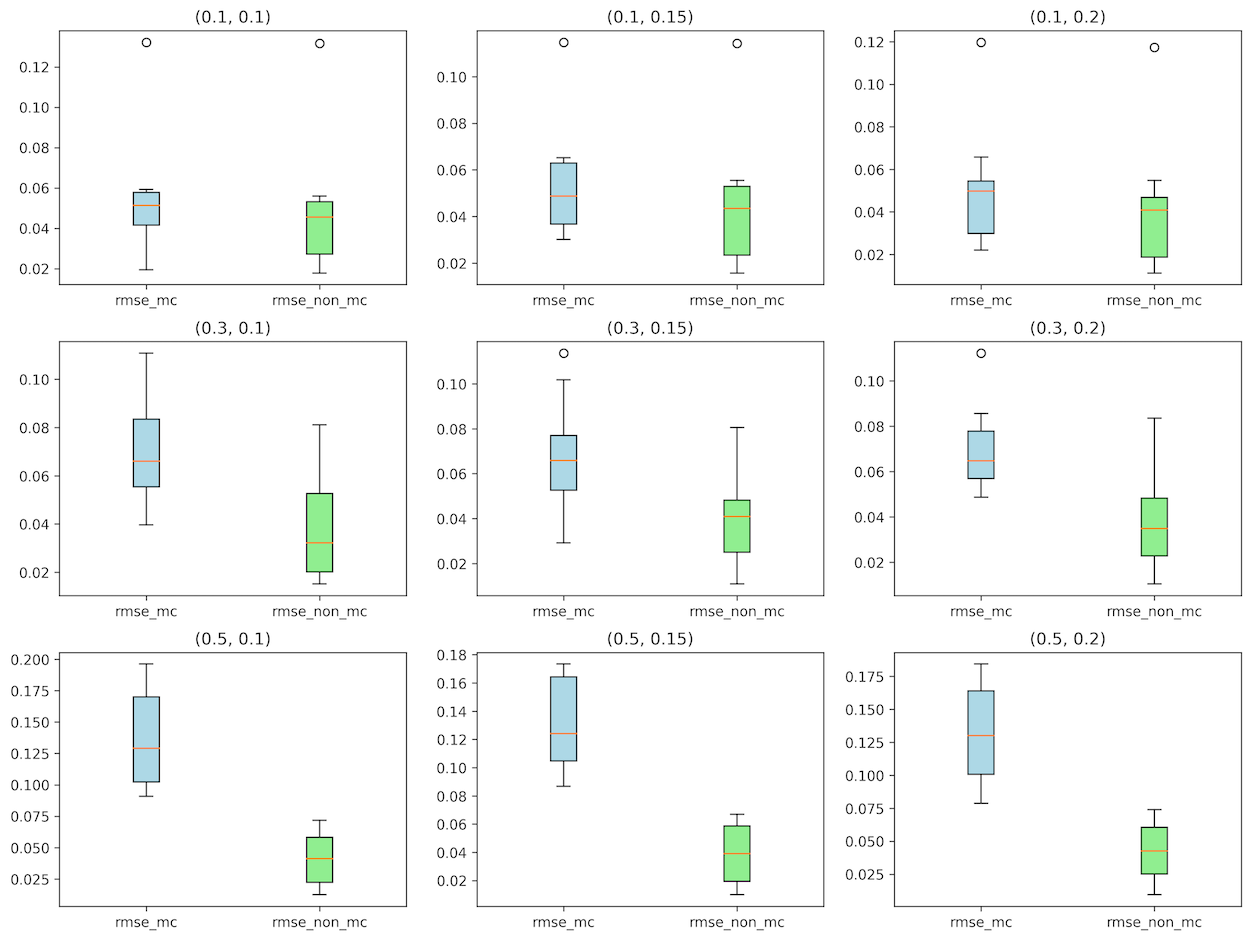}
\caption{3 layers, 4000 epochs}
\label{fig:1_0.01_3_4000box}
\end{subfigure}
\begin{subfigure}[t]{0.32\textwidth}
\includegraphics[width=\textwidth]{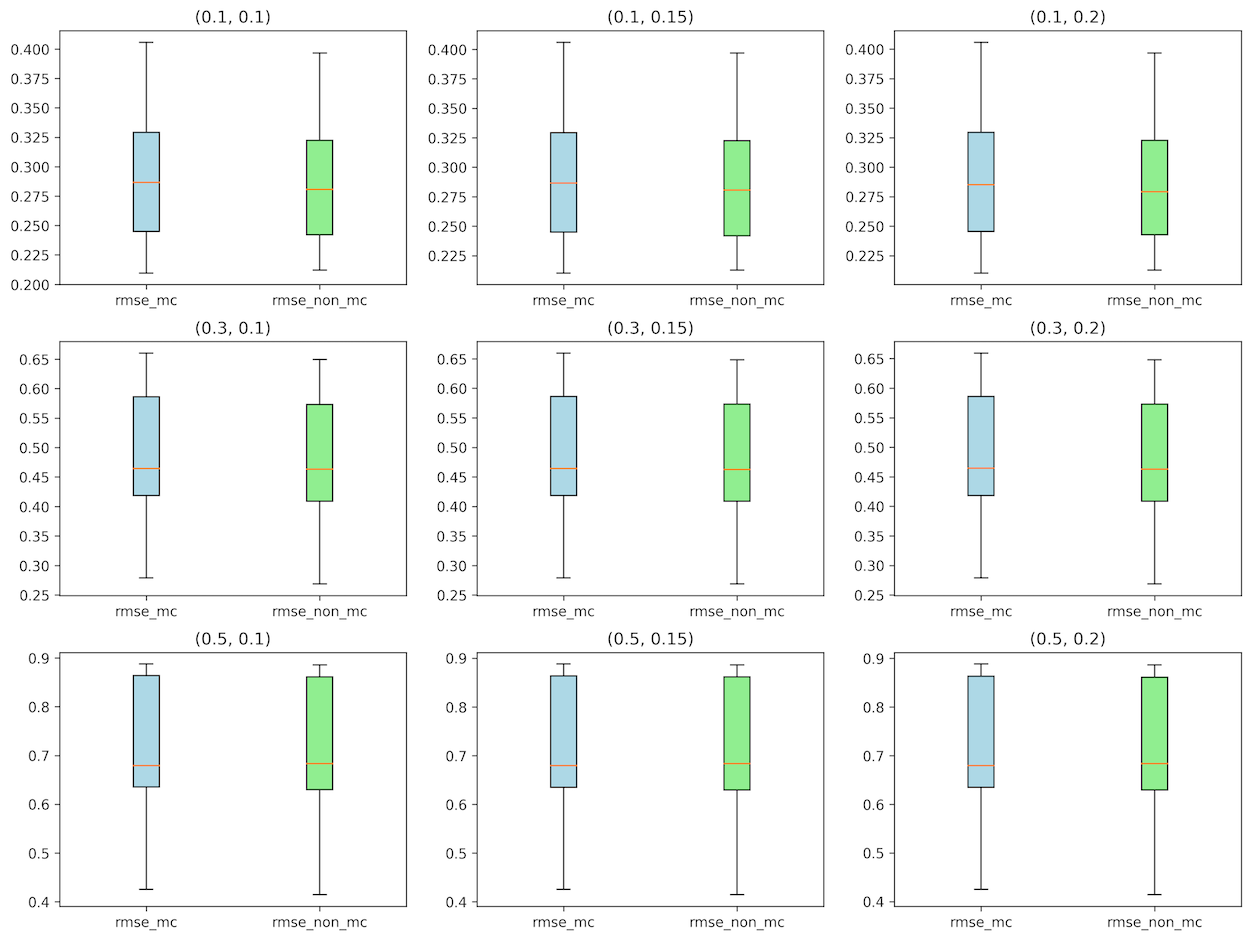}
\caption{5 layers, 40 epochs}
\label{fig:1_0.01_5_40box}
\end{subfigure}
\begin{subfigure}[t]{0.32\textwidth}
\includegraphics[width=\textwidth]{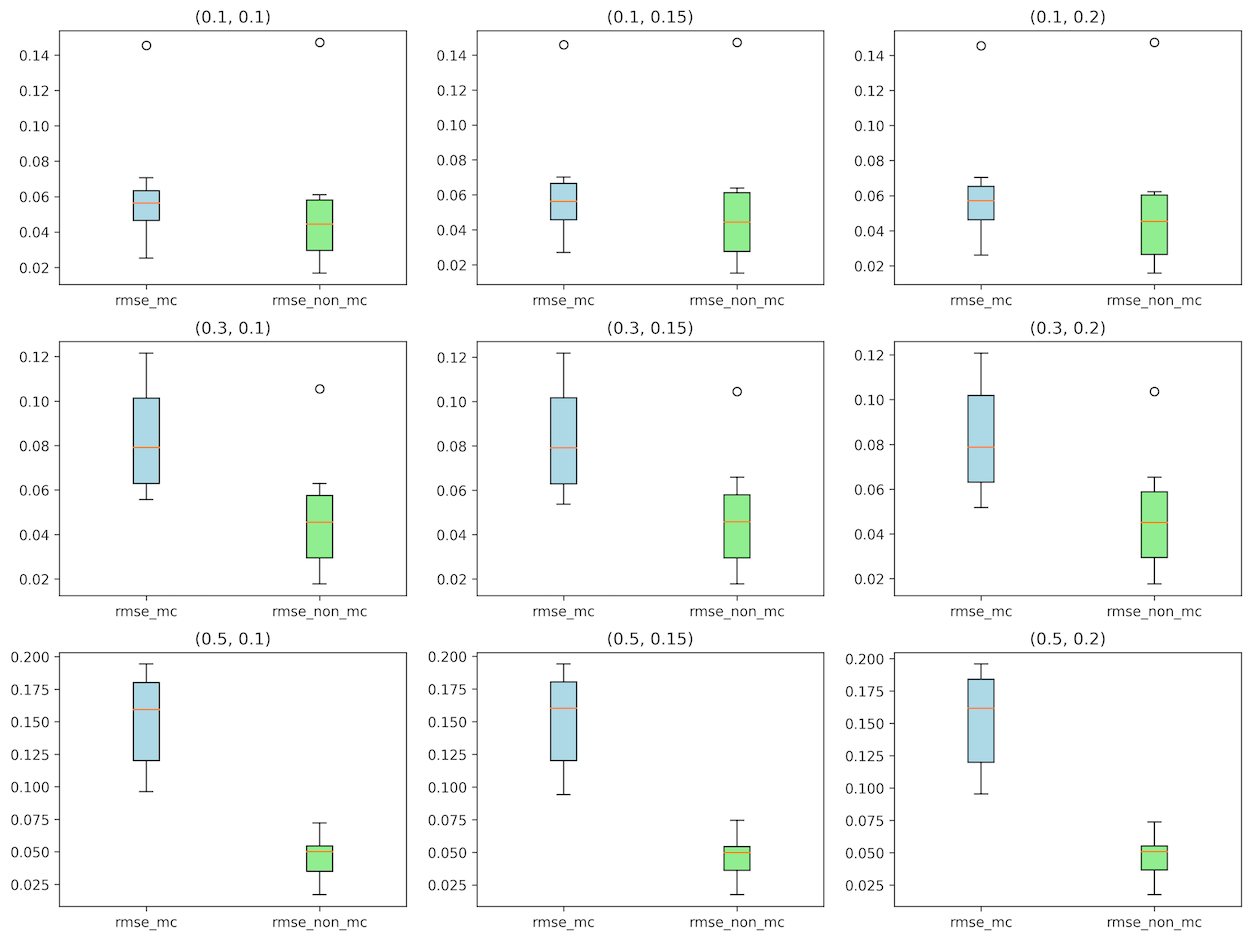}
\caption{5 layers, 400 epochs}
\label{fig:1_0.01_5_400box}
\end{subfigure}
\begin{subfigure}[t]{0.32\textwidth}
\includegraphics[width=\textwidth]{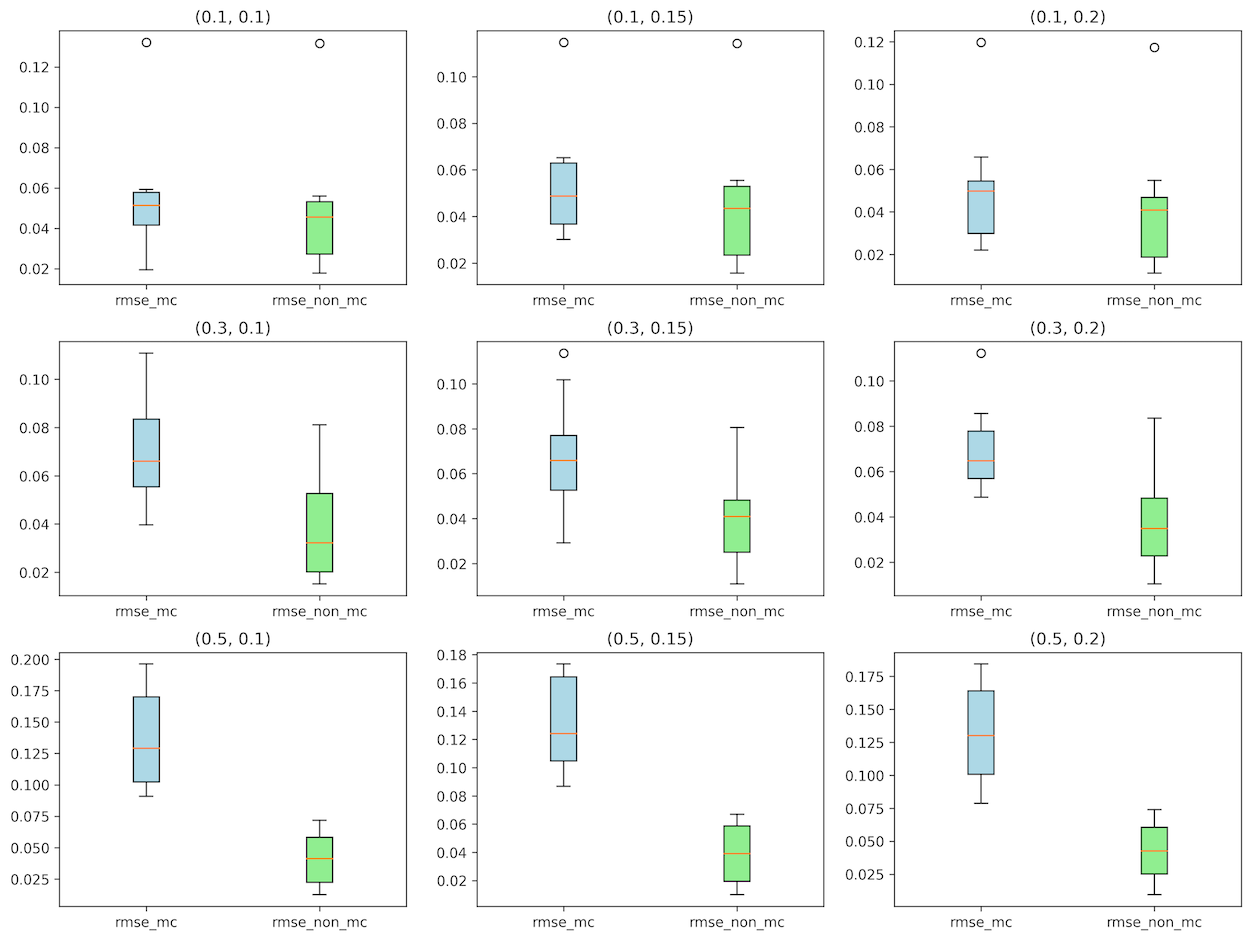}
\caption{5 layers, 4000 epochs}
\label{fig:1_0.01_5_4000box}
\end{subfigure}
\caption{Boxplots of RMSE scores across 10 different splits of \textsf{yacht} dataset on NNs with different number of layers, snapshot plots taken at 40, 400 and 4000 epochs. Each subplots \emph{within} the subfigure is for different dropout rates (0.1, 0.3, 0.5, y-axis) and \(\tau\) values (0.1, 0.15, 0.2) used. The blue boxplot is MC dropout, and the green boxplot is standard dropout.}
\label{fig:yacht_boxplot}
\end{figure}

We tried to examine the number of training iterations a little further, by trying to see if there's any relationship between the \emph{capacity} or strutural complexity of the network, and the number of training epochs. The following was the testing procedure:

\begin{itemize}
\item Using NNs of different number of hidden layers,
\item we trained them with different dropout rates (0.1, 0.3, 0.5, y-axis) and \(\tau\) values (0.1, 0.15, 0.2).
\item We calculate the score across 10 different train-test splits of the datasets to check the standard error. This was also done to see if MC dropout makes the scores across different splits more stable than the standard dropout.
\end{itemize}

The result is shown in \autoref{fig:yacht_boxplot}. The main conclusion from this plot was that the number of training iterations needed for the scores to converge do not have much to do with NN's structure. Even though we've tried different number of hidden layers, they all required similar number of training epochs to reach the same level of predictive performance. Dropout rates and \(\tau\) also did not seem to matter. We now suspect that the number of training iterations needed has more to do the size of the dataset and inherent aleatoric uncertainty within it.

\subsection{Number of Test Predictions}

As seen in \autoref{eq:mc_dropout}, we would need to make \(T\) test predictions and average them to get uncertainty estimates. Then how many test predictions should we make? \cite{srivastava2014dropout} claimed that around 50 predictions were needed for fully connected NNs on MNIST dataset, but it was done for networks where dropouts were added only before the final output layer.

\begin{figure}[h!]
\centering
\includegraphics[width=0.8\textwidth]{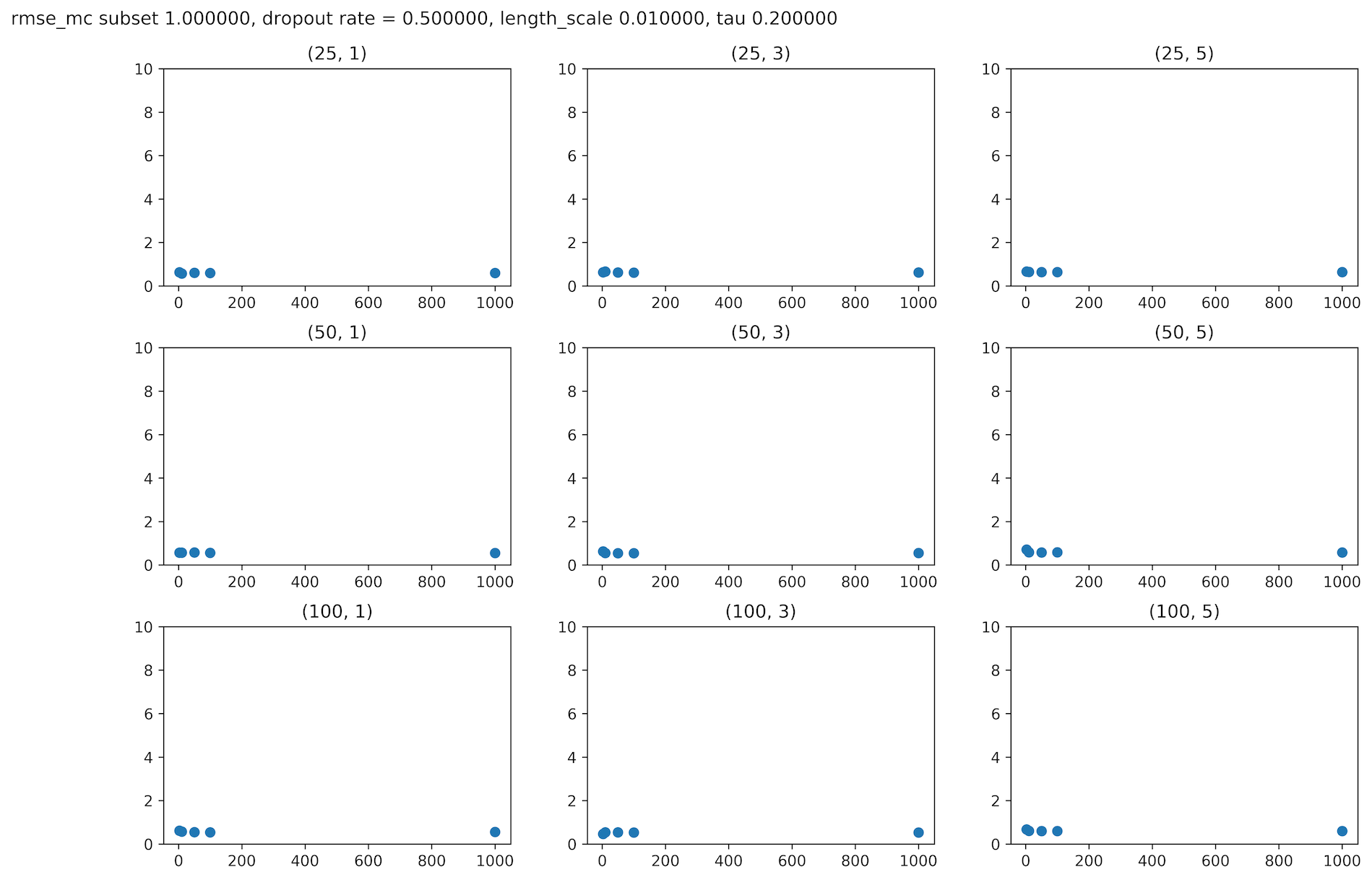}
\caption{RMSE scores obtained for different number of test predictions averaged, for \textsf{bostonHousing} dataset. Low values are better. 3, 10, 50, 100, and 1000 predictions from left to right. X-axis is for different number of layers, and y-axis tracks different layer sizes.}
\label{fig:test_predictions_boston}
\end{figure}

We can see in \autoref{fig:test_predictions_boston} that we only need to make very small number of predictions when it comes to predictive performance. While it is difficult to recommend an exact number, it seems that 50 is a safe choice.

\section{Conclusion / Future Directions}
Although many of the analyses in recent literatures, including the ones done in this paper, have focused on how it improves predictive performance, we find that the real value of MC dropout is that it provides a theoretically grounded and yet practical method for discovering uncertainty around our NN models. More importantly, it provides a potential for modelling uncertainties beyond epistemic model uncertainty.

\subsection{Aleatoric Uncertainty}
We mentioned previously that our current formulation of MC dropout assumes homoscedastic data noise. However, \cite{gal2016hetero}, \cite{gal2016uncertainty} and \cite{kendall2017uncertainties} have also shown that we could extend this to be heteroscedastic, esentially by having \(\tau\) to be a function of \(x\):

\begin{align}
\text{Var}_{q_\theta(y^\star \mid x^\star)}(y^\star) &\approx \frac{1}{T} \sum_{t=1}^{T} \tau(x^\star) + (f^{W}(x^\star))^T f^{W}(x^\star) \nonumber \\
&- (\mathbb{E}_{q_\theta(y^\star \mid x^\star)}(y^\star))^T\mathbb{E}_{q_\theta(y^\star \mid x^\star)}(y^\star) \label{eq:mc_dropout_var_hetero}
\end{align}

In practice, we make this \(\tau(x^\star)\) to be another output of our NN.

\begin{figure}[h!]
\centering
\includegraphics[width=0.5\textwidth]{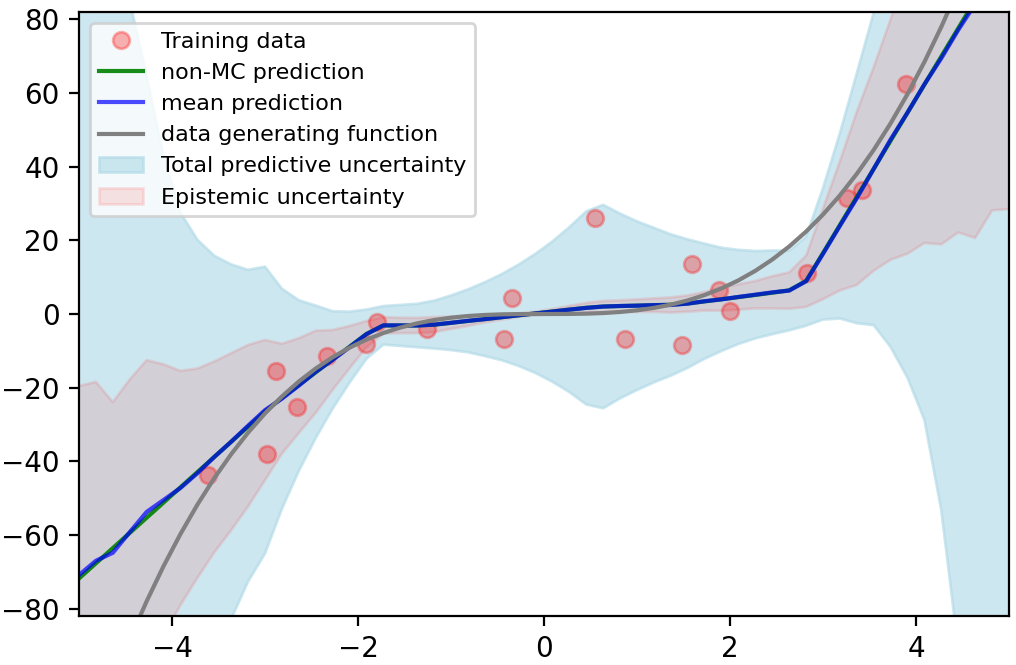}
\caption{Uncertainty captured from NN, now with heteroscedastic noise.}
\label{fig:hetero_aleatoric}
\end{figure}

We trained the same toy dataset used in \autoref{sec:uncertainty_info} with this alternative formulation. Now you can see that all points have been captured by total predictive uncertainty. However, it would be an very interesting discussion on whether this is a completely sound way of discerning epistemic and aleatoric uncertainty within the data. In our toy dataset, we injected the noise of \(\epsilon \sim \mathcal{N}(0,\,9)\). How much of that should be considered as epistemic uncertainty and other to be considered aleatoric?

Moreover, there have been some recent efforts that extends this approach and attempt to model aleatoric uncertainties arising in particular problem domains \cite{KWON2020106816}\cite{feng2019leveraging}\cite{tagasovska2018singlemodel}\cite{depeweg2019modeling}. Our future work will focus on analyzing these approaches and identifying implications for applications to other domains.

\appendix
\section{Program Codes}
All the program codes used for producing results presented in this paper are opensourced at \url{https://link.iamblogger.net/bdl-exp}.

{
\bibliographystyle{plain}
\bibliography{mc_dropout_qual_analysis}
}

\end{document}